\documentclass[10pt,twocolumn,letterpaper]{article}

\usepackage[pagenumbers]{iccv} % To force page numbers, e.g. for an arXiv version

\usepackage{times}
\usepackage{epsfig}
\usepackage{graphicx}
\usepackage{amsmath}
\usepackage{amssymb}
\usepackage{algorithm,algorithmic}
\usepackage{bm}
\usepackage{multirow}

\definecolor{iccvblue}{rgb}{0.21,0.49,0.74}
\usepackage[pagebackref,breaklinks,colorlinks,allcolors=iccvblue]{hyperref}

\def\paperID{2559} % *** Enter the Paper ID here
\def\confName{ICCV}
\def\confYear{2025}

\title{Crossmodal Knowledge Distillation with WordNet-Relaxed Text Embeddings for Robust Image Classification}

\author{
Chenqi Guo$^{1}$~~~~Mengshuo Rong$^{1}$~~~~Qianli Feng$^{2}$~~~~Rongfan Feng$^{1}$~~~~Yinglong Ma$^{1}$\\
$^1$North China Electric Power University \quad $^2$Amazon\\
{\tt\small \{chenqiguo72, mengshuorong\}@ncepu.edu.cn, fengq@amazon.com}\\
{\tt\small \{ronfangfeng, yinglongma\}@ncepu.edu.cn}
}

\begin{document}
\maketitle

\begin{abstract}

Crossmodal knowledge distillation (KD) aims to enhance a unimodal student using a multimodal teacher model. In particular, when the teacher’s modalities include the student’s, additional complementary information can be exploited to improve knowledge transfer. 
In supervised image classification, image datasets typically include class labels that represent high-level concepts, suggesting a natural avenue to incorporate textual cues for crossmodal KD.
However, these labels rarely capture the deeper semantic structures in real-world visuals and can lead to label leakage if used directly as inputs, ultimately limiting KD performance.
To address these issues, we propose a multi-teacher crossmodal KD framework that integrates CLIP image embeddings with learnable WordNet-relaxed text embeddings under a hierarchical loss. By avoiding direct use of exact class names and instead using semantically richer WordNet expansions, we mitigate label leakage and introduce more diverse textual cues.
Experiments show that this strategy significantly boosts student performance, whereas noisy or overly precise text embeddings hinder distillation efficiency. Interpretability analyses confirm that WordNet-relaxed prompts encourage heavier reliance on visual features over textual shortcuts, while still effectively incorporating the newly introduced textual cues.
Our method achieves state-of-the-art or second-best results on six public datasets, demonstrating its effectiveness in advancing crossmodal KD.

\end{abstract}

\section{Introduction}
\label{sec:intro}

\begin{figure}[h]
    \centering
    \includegraphics[width=0.9\columnwidth]{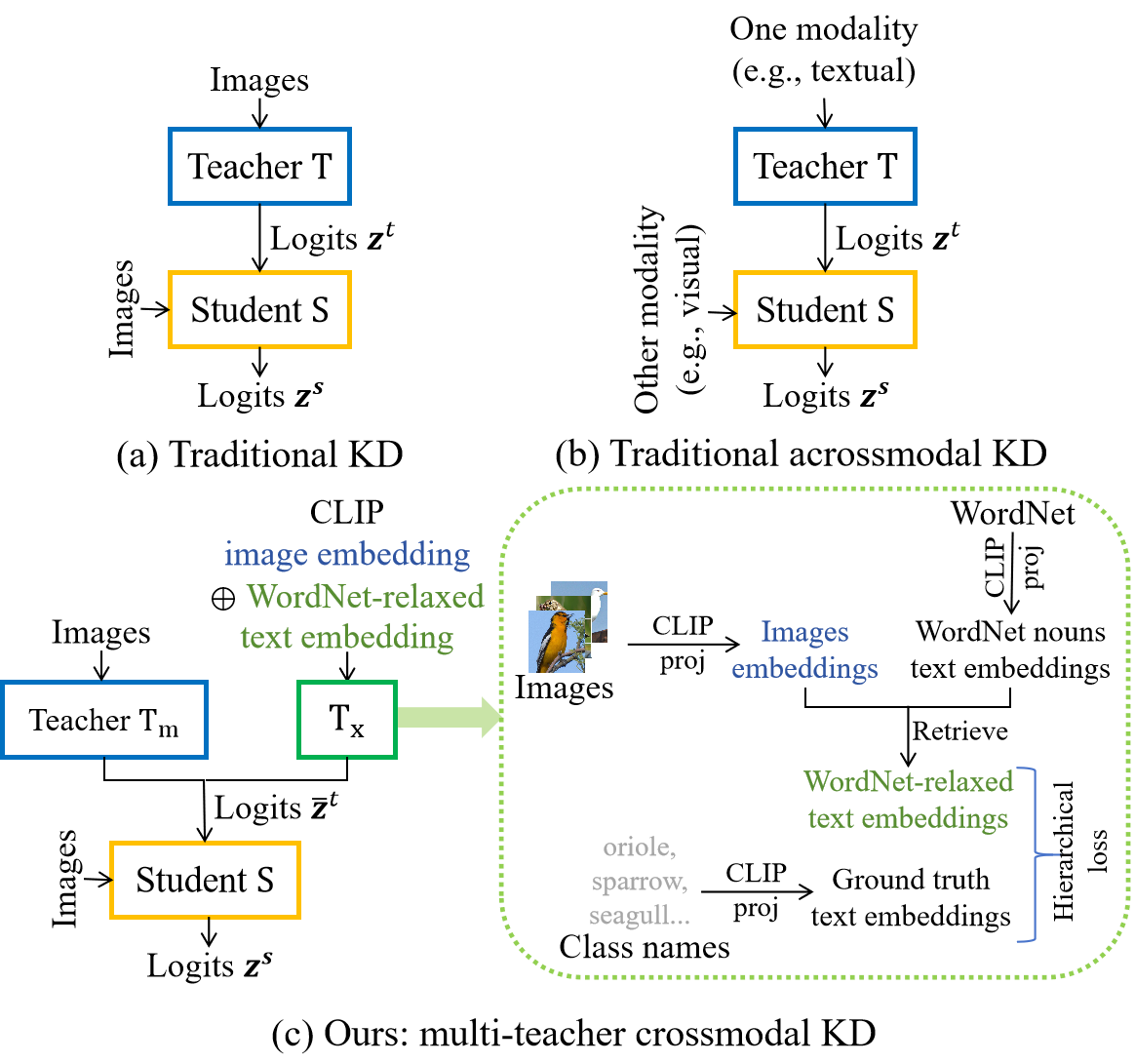}
    \vskip -0.1in
    \caption{Comparison of different KD frameworks for supervised image classification. (a) Vanilla KD: Teacher and student share the same modality (visual), providing limited training cues. (b) Conventional acrossmodal KD: Knowledge is transferred from one modality (e.g., text) in the teacher to another modality (e.g., image) in the student, but suffers from modality gaps and insufficient general modality features in the teacher. (c) Our multi-teacher crossmodal KD with WordNet-relaxation: Combines both image and CLIP-based multimodal embeddings in the teacher, providing richer information to the student. 
    }
    \label{fig:method_comparison}
    \vskip -0.2in
\end{figure}

Knowledge distillation (KD) \cite{hinton2015distilling} has primarily focused on unimodal settings, where both teacher and student share the same visual modality. Progress in \emph{crossmodal} KD \cite{Gupta7780678}, however, leverages multimodal teacher models to enhance a unimodal student. Broadly, crossmodal KD can involve transferring knowledge from a teacher’s modality to a different student modality, or using a teacher whose modalities \emph{include} the student’s, thus enriching the student’s existing modality with complementary information. In this work, we focus on the latter scenario, where the unimodal student benefits from additional cues (e.g., text) while retaining its primary vision modality, enabling more robust learning in supervised image classification.

Despite these benefits, previous research \cite{xue2023modality} has raised concerns about the efficacy of crossmodal KD. In particular, it has been demonstrated that a more accurate teacher with additional modalities does not always improve unimodal student performance. 
Their concept of ``modality-general decisive features'', as defined by the MVD (modality Venn diagram) generation rule, suggests that boosting the overlap of shared features between the teacher modality $a$ and student modality $b$ is critical for enhancing knowledge transfer. In this work, we adapt this principle to our $(a,b)\rightarrow b$ scenario, where the teacher employs both text and image, yet the student relies solely on images, by focusing on ``general modality features'' that emphasize the shared visual domain.

Nevertheless, putting this principle into practice requires a strategy for effectively leveraging both the teacher’s and student’s modalities. Intuitively, image datasets inherently come with class labels that capture high-level concepts, suggesting a natural opportunity to incorporate textual information for crossmodal KD. While these labels provide baseline supervision, they often fail to capture the complex semantic nuances present in real-world visuals. We hypothesize that augmenting class names with semantically expanded cues rather than simply relying on their exact forms, can complement the visual data with additional, more diverse signals for crossmodal KD.

Yet, implementing such textual augmentation for KD requires careful handling of label semantics. While crossmodal few-shot learning \cite{multihelpuni2023} offers insights into leveraging CLIP \cite{CLIP2021} text embeddings as virtual samples, our experiments confirm that naively relying on ground-truth class-name prompts causes severe label leakage, and merely injecting noise fails to fully leverage the additional textual modality. In both cases, either the teacher’s reliance on the general visual modality features is undercut by textual shortcuts, or semantic coherence is compromised by random noise, ultimately degrading student performance. These observations highlight the need for a more nuanced strategy that maintains semantic integrity, ensuring the teacher both increases reliance on the general visual modality and effectively incorporates textual cues in crossmodal KD.

This observation motivates us to leverage WordNet \cite{wordnet1995}, a lexical database that organizes words into semantically related groups, to enrich CLIP text embeddings. Drawing inspiration from zero-shot learning methods that incorporate WordNet cues \cite{ge2023improving}, and TAC \cite{li2024image} which boosts clustering by retrieving discriminative WordNet nouns, we adapt these expansions for supervised KD in image classification. Unlike TAC, which uses fixed WordNet expansions for clustering, our \emph{WordNet-relaxed} embeddings act as learnable parameters in the teacher model, serving as a semantic regularizer. This approach reduces reliance on exact class labels, mitigates label leakage, strengthens the teacher’s general modality features, and more effectively leverages the additional textual cues. To ensure deeper semantic alignment between the WordNet-relaxed and exact class-name-based text embeddings, we introduce a hierarchical loss which compares WordNet-relaxed and exact class-name embeddings, and a cosine regularization which prevents excessive drift from pretrained distributions, resulting in a more robust framework.

Building on the training strategy from \cite{Omnivore2022}, which trains a single model to predict labels across various classification tasks using multiple modalities with identical parameters, we propose a multi-teacher crossmodal KD framework that uses WordNet-relaxed text embeddings (see Figure~\ref{fig:method_comparison}\,(c)). 
In this setup, a unimodal teacher $\text{T}_m$ can encompass an ensemble of different image-based teachers (e.g., various data augmentations \cite{GUO2025125579}). Meanwhile, a multimodal teacher $\text{T}_x$ processes both the student’s vision modality and an additional WordNet-relaxed text modality, with all embeddings generated via CLIP. 
By leveraging these complementary teachers, one unimodal and one multimodal, the student gains a more comprehensive perspective, ultimately improving performance in supervised image classification.

Our contributions are fourfold:
\begin{enumerate}
    \item We propose a multi-teacher crossmodal KD pipeline that incorporates one unimodal teacher focusing on image augmentations, and another multimodal teacher employing CLIP image and text embeddings.
    \item We introduce a semantic regularization mechanism that encourages the teacher model to learn from more generalized, ``relaxed'' textual descriptors.
    Experiments are conducted to analyze how varying proportions of WordNet-relaxed text (vs.\ exact or noisy class names) influence teacher and, in turn, affect student accuracy.
    \item We demonstrate, through interpretability analyses using Captum, that our WordNet-based regularization drives the teacher model to rely more on robust general visual modality features, reduces textual memorization, and effectively utilizes the additional textual cues introduced.
    \item We conduct extensive experiments across six public datasets, and results show that our method achieves SOTA or, at the very least, the second-best performance.
\end{enumerate}

\section{Related Work}
\label{sec:RelatedWork}

\noindent\textbf{CLIP KD and Crossmodal KD:}
CLIP Knowledge Distillation (KD) efforts primarily compress large CLIP models into smaller ones \cite{yang2024clip} or distill vision-language knowledge for tasks like Visual Question Answering \cite{CLIPTD2022}.
C$^2$KD \cite{Huo2024CVPR} addresses crossmodal KD by transferring knowledge from one teacher modality to a different student modality, employing bidirectional distillation and selective filtering of misaligned soft labels to bridge modality gaps.
Another crossmodal KD variant, particularly relevant to our work, uses a \emph{multimodal teacher} that processes both the student’s modality (e.g., vision) and an additional modality (e.g., text), providing complementary cues.
Although \cite{xue2023modality} underscores the importance of overlapping ``modality-general decisive features'' between teacher and student, we focus on the scenario where the teacher incorporates both image and text modalities for a unimodal image student.
To this end, we feed CLIP image embeddings and \emph{WordNet-relaxed} text embeddings into the teacher. Our semantic regularizer, which replaces class names with WordNet synonyms or hypernyms, mitigates label leakage, enriches crossmodal supervision, boosts general modality features, and utilizes textual cues more effectively in the teacher, thereby boosting knowledge transfer for unimodal image classification.

\noindent\textbf{Multimodal Few-shot or Zero-shot Learning:} 
Multimodal pre-trained models like CLIP often excel at few-shot or zero-shot classification by leveraging text-image alignment. For instance, \cite{transduct2024} fuses visual and textual features for transductive few-shot classification, while \cite{ge2023improving} augments zero-shot CLIP prompts with hierarchical labels to handle uncertainty. TAC \cite{li2024image} uses WordNet nouns for clustering by selecting discriminative terms and distilling crossmodal neighborhood information. Our approach likewise employs WordNet expansions but for supervised KD in image classification, not clustering or few-shot learning. We further propose a Hierarchical Loss (comparing ground-truth and WordNet-relaxed embeddings) and a Cosine Regularization term (comparing pretrained and WordNet-relaxed embeddings) to ensure deeper semantic alignment, yielding a more robust KD framework.

\section{Problem Definition}\label{sec:problemDef}

We focus on multi-teacher crossmodal knowledge distillation (KD) for supervised image classification, where a smaller student model $\text{S}$ learns from the logits of multiple teacher models $\text{T}_i$. Unlike traditional unimodal KD, our approach leverages multimodal cues by incorporating CLIP-based image embeddings and/or WordNet-relaxed text embeddings into one of the teachers. The combined logits from the teachers are then distilled into the unimodal (image-only) student. In this section, we outline the main challenges and explain how our framework addresses them.

\noindent\textbf{Multi-Teacher Setup:}
We employ two types of teachers to provide the student with a more comprehensive perspective: (1) \textbf{Unimodal Teacher} $\text{T}_m$: An ensemble of one or more image-based teachers (e.g., with diverse data augmentations \cite{GUO2025125579}) operating solely on the student’s vision modality. (2) \textbf{Multimodal Teacher} $\text{T}_x$: A classifier that processes both the student’s vision modality and an additional text modality.
Each teacher produces logits, which are combined into a single distribution for the student to match. See Section~\ref{sec:ensemKD} for further details on combining teachers’ outputs and training the student.

\noindent\textbf{Label Leakage vs. Textual Regularization:}
We hypothesize, and later empirically confirm, that naively using exact class-name-based prompts in text embeddings to a classifier leads to ``deception'' (label leakage). This results in artificially high teacher accuracy that fails to provide meaningful KD signals. 
To address this, we propose using WordNet expansions and regularizers for textual prompts to enrich semantics without directly providing class tokens. This approach mitigates textual memorization while enhancing the contribution of general modality features, i.e., the visual modality, while also effectively leverages the newly introduced textual features. Detailed explanations are provided in Section \ref{sec:wordnetrelax}.

\noindent\textbf{Semantic Alignment:}
We introduce hierarchical loss and cosine regularization to ensure that the relaxed text embeddings remain aligned with the true class space, preserving the meaningful class structure. Further details are provided in Section \ref{sec:hierloss}.

\section{Methodology}

Our approach leverages WordNet-based textual expansions to enrich representation learning, but instead of focusing on unsupervised clustering, we target knowledge distillation (KD). We introduce a hierarchical loss to ensure richer semantic text prompts and deeper semantic alignment for the input multimodal features to the teacher. This approach regularizes the text embeddings to capture broader category nuances, reinforcing the teacher’s reliance on the general visual modality features while effectively incorporating textual cues, ultimately yielding a more robust crossmodal KD framework.

\subsection{Multi-teacher KD with Multimodal Inputs}\label{sec:ensemKD}

\noindent\textbf{Teachers:}
We employ two types of teachers with distinct input modalities.
The unimodal teacher \textbf{$\text{T}_m$} is a state-of-the-art ensemble of image-based classifiers \cite{GUO2025125579} that use images augmented via different strategies: strong policy (e.g., RandAugment \cite{RandAugment2020} for ImageNet or AutoAugment \cite{AutoAugment2019} for other datasets), and weak policy (e.g., random cropping, horizontal flipping, and mild color jitter).
In contrast, the multimodal teacher \textbf{$\text{T}_x$} is a classifier processing CLIP image embeddings concatenated with WordNet-relaxed text embeddings, where class names are expanded or partially replaced with semantically related nouns from WordNet. This approach avoids direct reliance on exact class names, thereby mitigating label leakage and promoting robust semantic alignment.

Given a classification task with input space $\mathcal{X}=\{\bm{x}_i\}^N_{i=1}$ and label space $\mathcal{Y}=\{y_c\}^C_{c=1}$, our teachers output logits $\{\mathbf{z}^{tm}, \mathbf{z}^{tx}\}$. We denote $\overline{\mathbf{z}}^t$ as the average of teacher logits:
\begin{equation}
    \overline{\mathbf{z}}^t = \frac{\mathbf{z}^{tm} + \mathbf{z}^{tx}}{2}
\end{equation}

\noindent\textbf{Student:}
A smaller classifier $\text{S}$ is trained from scratch, producing logits $\mathbf{z}^s$. Denoting $\sigma_c(\cdot)$ as the softmax function, the standard KD objective with temperature $\tau$ becomes:
\begin{equation}
    \mathcal{L}_\mathrm{KD}(\mathbf{z}^s, \overline{\mathbf{z}}^t) = -\tau^2 \sum_{c=1}^C \sigma_c\!\Bigl(\frac{\overline{\mathbf{z}}^t}{\tau}\Bigr)_{c} \log \sigma_c\!\Bigl(\frac{\mathbf{z}^s}{\tau}\Bigr)_{c}
\end{equation}

We also include the usual cross-entropy with the ground-truth label $y$, so the total objective is:
\begin{equation}
    \mathcal{L} = \underbrace{-\sum_{c=1}^C y_c \log \sigma_c(\mathbf{z}^s)_c}_\text{supervised CE} \;+\;\lambda \,\mathcal{L}_\mathrm{KD}\bigl(\mathbf{z}^s, \overline{\mathbf{z}}^t\bigr)
\end{equation}

In summary, the final student model operates solely on images during inference, indirectly benefiting from the teacher's multimodal knowledge while avoiding direct textual leakage.

\subsection{WordNet-Relaxed Textual Embeddings}\label{sec:wordnetrelax}

Our experiments reveal label leakage when using ground-truth class names as prompts (e.g., ``A photo of a telephone'') for CLIP text embeddings. While this results in artificially high teacher accuracy, it provides limited benefits for KD, as the teacher is deceptive through memorizing the textual cue. 
Conversely, randomly shuffling class names introduces label noise that diminishes this ``cheating'' and enhances the contribution of general modality features in the teacher model, yet fails to exploit the additional textual cues effectively, ultimately degrading student performance compared to WordNet-relaxation.
Further evidence can be found in Section \ref{sec:ablation_noise} and \ref{sec:interpretability_tsne}.

\noindent\textbf{WordNet-Relaxation Text Generation:}
To mitigate label leakage and effectively leverage text embeddings, we propose using WordNet-relaxed prompts. Instead of relying on an exact class name (e.g., ``telephone''), we retrieve semantically related synonyms or hypernyms (e.g., ``desk phone'', ``telephone booth'', etc.) from WordNet for each image sample. This broadens the semantic scope of the prompt without providing a direct class token. The procedure is outlined in Algorithm~\ref{alg:wordnet}.

\begin{algorithm}[t]
\caption{\textbf{WordNet-Relaxation Text Generation}}
\label{alg:wordnet}
\begin{algorithmic}[1]
\renewcommand{\algorithmicrequire}{\textbf{Input:}}
\renewcommand{\algorithmicensure}{\textbf{Output:}}

\REQUIRE 
\begin{itemize}
\item A set of candidate WordNet nouns $\{n_i\}$;
\item A CLIP text encoder $\mathrm{encodeText}(\cdot)$ and a set of prompt templates $\{\mathcal{T}_k\}$;
\item A CLIP image encoder $\mathrm{encodeImage}(\cdot)$;
\item Hyperparameters: batch size $b$, cluster number $M$.
\end{itemize}

\ENSURE A filtered and normalized set of WordNet-relaxed noun embeddings $\{\mathbf{n}^\star_j\}$ for subsequent multimodal teacher training.

\vspace{1mm}
\hrule
\vspace{1mm}

\STATE \textbf{Step 1: Infer Noun Embeddings.}
\STATE Initialize an empty array \texttt{features} to store embeddings.
\FOR{each template $\mathcal{T}_k \in \{\mathcal{T}_k\}$}
  \FOR{each mini-batch of noun strings $\{n_i\}$ of size $b$}
    \STATE Construct prompts: $\text{prompt}_i = \mathcal{T}_k(n_i)$
    \STATE $\mathbf{f}_{\mathrm{text}} \gets \mathrm{encodeText}(\text{prompt}_i)$
    \STATE Append $\mathbf{f}_{\mathrm{text}}$ to \texttt{features}
  \ENDFOR
\ENDFOR
\STATE Average:
$
\mathbf{n}_i \leftarrow 
\frac{1}{|\{\mathcal{T}_k\}|}
\sum_{k} \mathbf{f}_{\mathrm{text}, i},
\quad 
\mathbf{n}_i \leftarrow 
\frac{\mathbf{n}_i}{\|\mathbf{n}_i\|}.
$

\vspace{1mm}
\hrule
\vspace{1mm}

\STATE \textbf{Step 2: Filter via K-Means Alignment.}
\STATE Compute CLIP image embeddings: \\
$\mathbf{v}_j = \mathrm{encodeImage}(\mathbf{x}_j)$ for all training images $\mathbf{x}_j \in \mathcal{D}_\mathrm{train}$.
\STATE Cluster $\{\mathbf{v}_j\}$ into $M$ centers $\{\mathbf{m}_k\}$ using K-means.
\STATE Compute similarity matrix 
$\mathbf{S} \in \mathbb{R}^{M \times \#(\mathbf{n}_i)}$ where 
$\mathbf{S}_{k,i} = \mathbf{m}_k \cdot \mathbf{n}_i$.
\STATE Assign each noun embedding $\mathbf{n}_i$ to the cluster $k = \arg\max_{k} \mathbf{S}_{k,i}$.
\STATE For each cluster, select top-5 noun embeddings by highest softmax score 
$\mathrm{softmax}(\mathbf{S}_{k,i})$.

\vspace{1mm}
\hrule
\vspace{1mm}

\STATE \textbf{Step 3: Finalize \& Update Noun Embeddings.}
\STATE Concatenate selected noun indices from each cluster to form $\{\mathbf{n}^\star_j\}$.
\STATE Re-normalize each $\mathbf{n}^\star_j \leftarrow \mathbf{n}^\star_j / \|\mathbf{n}^\star_j\|$.
\STATE \textbf{Set $\{\mathbf{n}^\star_j\}$ as learnable parameters}, updated during the multimodal teacher training via hierarchical loss and cosine regularization.

\RETURN $\{\mathbf{n}^\star_j\}$

\end{algorithmic}
\end{algorithm}

\noindent\textbf{Handling the Final Embeddings.} 
The filtered embeddings $\{\mathbf{n}^\star_j\}$ represent a curated subset of synonyms/hypernyms that more faithfully span the semantic neighborhood of each class while removing outliers. We then treat each $\mathbf{n}^\star_j$ as a trainable parameter and load these vectors into the teacher’s training loop, where they can be fine-tuned with the image embeddings under semantic constraints. 
See Section \ref{sec:hierloss} for how we further update these embeddings via hierarchical loss and cosine regularization. By broadening textual descriptions and removing direct label tokens, we can preserve enough textual variety to enrich the teacher’s multimodal representation, significantly mitigating label leakage, maintaining coherent text semantics, and fully leveraging textual cues while reinforcing the teacher’s reliance on general visual modality features.

\subsection{Hierarchical Loss and Cosine Regularization}\label{sec:hierloss}

While WordNet expansions help enrich the text embeddings, updating the selected noun embeddings allows the model to adapt these representations to the specific task, ensuring better alignment with the target classes.
However, this updating process risks drifting too far from the true class semantics or the pretrained embeddings. Hence, we introduce two regularizers to maintain alignment with the original semantics while allowing flexibility for adaptation.

\noindent\textbf{Hierarchical Loss:}
Let $\mathbf{n}_\mathrm{gt}$ be the text embedding for the exact class name, used only for semantic alignment.
Meanwhile, $\mathbf{n}_\mathrm{relaxed}$ is the updated WordNet-relaxed text embedding from $\{\mathbf{n}^\star_j\}$ corresponding to each image embedding during training.
We define:
\begin{equation}
    \mathcal{L}_\mathrm{hier} 
   = 1 - \cos~\!\bigl(\mathbf{n}_\mathrm{gt}, \,\mathbf{n}_\mathrm{relaxed}\bigr)
\end{equation}
ensuring that $\mathbf{n}_\mathrm{relaxed}$ remains consistent with the real class descriptor while still capturing broader synonyms.

\noindent\textbf{Cosine Regularization:}
To prevent the learned WordNet-relaxed text embeddings from straying excessively from their pretrained WordNet distribution, we penalize large deviations:
\begin{equation}
    \mathcal{L}_\mathrm{cosreg}
   = 1 - \cos~\!\bigl(\mathbf{n}_\mathrm{pretrained}, \,\mathbf{n}_\mathrm{relaxed}\bigr)
\end{equation}
where $\mathbf{n}_\mathrm{pretrained}$ is the original reference to $\mathbf{n}^\star_j$ before any updates. This helps keep the textual perturbations reasonable and stabilizes the teacher's knowledge.

\noindent\textbf{Overall Teacher Objective:}
Teacher model $\text{T}_x$ thus optimizes:
\begin{equation}
    \mathcal{L}_{\text{T}_x} = \mathcal{L}_\mathrm{sup}(\mathbf{z}^{tx}, y) + \lambda_\mathrm{hier} \,\mathcal{L}_\mathrm{hier}  + \lambda_\mathrm{cosreg} \,\mathcal{L}_\mathrm{cosreg}
\end{equation}
with $\lambda_\mathrm{hier},\lambda_\mathrm{cosreg}$ controlling the strength of hierarchical loss and cosine regularization. Here we set $\lambda_\mathrm{hier}=0.1$ and $\lambda_\mathrm{cosreg}=0.01$.

\section{Experimental Setup}

We evaluate our framework across standard classification benchmarks, focusing on ImageNet \cite{deng2009imagenet} and its long-tailed variant ImageNet-LT, CIFAR100 \cite{CIFAR2009} and its imbalanced version CIFAR100-imb100, as well as the Scene \cite{sceneDatasetKaggle} and UTKFace \cite{UTKFaceDatasetKaggle} datasets.

For the unimodal teacher \textbf{$\text{T}_m$}, we utilize an ensemble of two teachers $\text{T}_{1s}$ and $\text{T}_{2w}$, which are two ResNet50 classifiers pretrained on ImageNet and fine-tuned on images from their respective target datasets using strong and weak data augmentations, respectively, as detailed in Section \ref{sec:ensemKD}. 
For the multimodal teacher \textbf{$\text{T}_x$}, it is also a pretrained ResNet50 classifier, but fine-tuned on CLIP image and/or text embeddings, with additional regularization applied through $\mathcal{L}_\mathrm{hier}$ and $\mathcal{L}_\mathrm{cosreg}$. 
During inference, teacher models are evaluated using their respective input types.
As a result, the validation accuracies of the teachers are not directly comparable, though their performance is presented in Table \ref{tab:TS_concat_clips__ablation_TS_noise}.

The student model is ResNet18 trained from scratch using strong data augmentation (i.e., $\text{S}_s$).
Notably, at inference time, the final student model uses only images for classification, with no reliance on text or image embeddings, thereby demonstrating the effectiveness of utilizing multimodal cues exclusively during training.

During training, we control the use of text embeddings by varying the proportions of WordNet expansions or noised class names in prompts. Through this setup, we evaluate how the proposed multi-teacher crossmodal KD enhances the student's classification performance. 
%More details on the hyperparameters are provided in supplementary material Section 1.

\section{Experimental Results and Analysis}

\begin{table*}[h]
\begin{center}
\begin{small}
\begin{tabular}{lcccccc}
\hline
\multirow{2}{*}{\textbf{Dataset}} 
& \multicolumn{6}{c}{\textbf{Student Top-1 Val Accuracy (\%)}} \\ \cline{2-7}
& LFME & DMAE & FFKD & Z-Score & Baseline$^{1}$ & Ours$^{2}$ \\
\hline
ImageNet & - & \textbf{81.98} & \underline{70.17} & 67.81 & 68.95 & \underline{70.14} \\
ImageNet-LT & 38.80 & 43.95 & 46.84 & 48.05 & \underline{49.65} & \textbf{50.06} \\
CIFAR100 & - & 79.60 & 79.80 & 80.44 & \underline{82.04} & \textbf{83.33} \\
CIFAR100-imb100 & 43.80 & 37.25 & 50.01 & 52.19 & \textbf{52.87} & \underline{52.58} \\
Scene & - & 88.62 & 88.27 & \underline{92.55} & 92.13 & \textbf{93.07} \\
UTKFace & - & 79.85 & 78.15 & \underline{86.50} & 85.24 & \textbf{86.71} \\
\hline
\multicolumn{7}{p{310pt}}{$^1$ $\text{T}_{1s}\text{T}_{2w}\text{T}_{3s}\text{S}_s$} \\
\multicolumn{7}{p{310pt}}{$^2$ $\text{T}_m\text{T}_{x,~(\text{img},~0\%\text{gt}~100\%\text{wn})}\text{S}_s$, where $\text{T}_m$ is the teacher ensemble $\text{T}_{1s}\text{T}_{2w}$} \\
\end{tabular}
\end{small}
\end{center}
\vskip -0.2in
\caption{Comparison of the final student performance across six datasets. The best and second-best results are highlighted in bold and underlined, respectively. LFME is specifically tailored for long-tailed datasets, so we only present its results on those.}
\label{tab:final_results}
\vskip -0.1in
\end{table*}

\subsection{Quantitative Evaluation}\label{sec:final_comparison}

This section presents our results across six datasets: ImageNet, ImageNet-LT, CIFAR100, CIFAR100-imb100, Scene, and UTKFace. Table~\ref{tab:final_results} shows the student’s top-1 validation accuracy for our best-performing trial ($\text{T}_m\text{T}_{x,~(\text{img},~0\%\text{gt}~100\%\text{wn})}\text{S}_s$) compared to the baseline ensemble KD (i.e., three-teacher ensemble KD $\text{T}_{1s}\text{T}_{2w}\text{T}_{3s}\text{S}_s$), and other SOTA supervised KD classifiers, including LFME \cite{LFME2020}, DMAE \cite{DMAE2023}, FFKD \citep{FFKD2024}, and Z-score logit standardization KD \cite{Sun2024Logit}. 
For a fair comparison, all KD methods use the same classifier backbones: ResNet50 for teachers and ResNet18 for students, except DMAE, which employs a ViT-Base-patch16 architecture for both. Note that LFME is specifically designed for long-tailed datasets, so its results are only shown in those scenarios. 
In every case, our approach achieves either new SOTA results or, at the very least, the second-best performance.
It is important to note that while DMAE achieves very high accuracy on ImageNet, its ViT architecture requires significantly more computational resources and running time compared to ours. Additionally, the ViT nature of DMAE is more suitable for large-scale datasets like ImageNet, highlighting the efficiency of our method for diverse dataset sizes.

Overall, our method demonstrates that combining WordNet-relaxed text prompts with CLIP image embeddings in a multi-teacher framework yields SOTA accuracy across diverse benchmarks.
Having established the method’s effectiveness, we next dissect its key components: WordNet relaxation, concatenating CLIP image and text embeddings, ablation with noisy text prompts or learnable WordNet text embeddings, and interpretability analysis, to further illuminate each element’s contribution.

\begin{figure}[h]
    \centering
    \includegraphics[width=0.9\columnwidth]{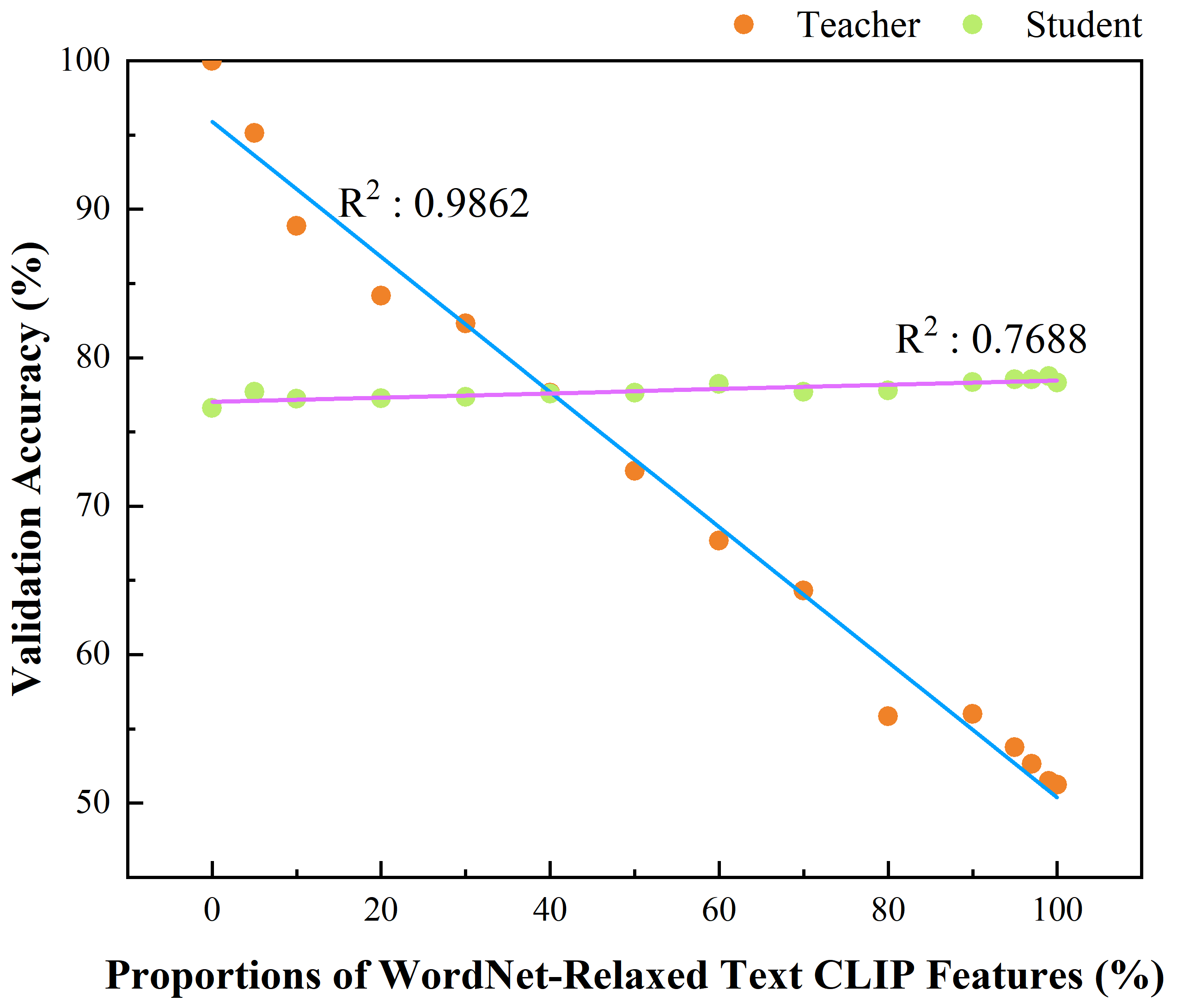}
    \vskip -0.1in
    \caption{Teacher $\text{T}_x$ and Student $\text{S}_s$ top-1 validation accuracy under KD on CIFAR100, across varying proportions of CLIP WordNet-relaxed text embeddings. Note that $\text{T}_{x}$ uses only CLIP text embeddings as inputs here. The proportion refers to the ratio of training samples using WordNet-relaxed text embeddings versus those using ground truth class-name-based ones. As the proportion increases, the teacher’s validation accuracy decreases, indicating that the classification task becomes more challenging. In contrast, the student performance improves with a higher proportion of WordNet-relaxed text embeddings, highlighting the regularization benefits of incorporating more diverse semantic cues.}
    \label{fig:wordnet_performance}
    \vskip -0.2in
\end{figure}

\subsection{Investigation: Impact of CLIP WordNet-Relaxed Text Embeddings}\label{sec:wordnettext}

In this experiment, we explore the effect of using CLIP text embeddings based on ground truth class names versus WordNet-relaxed text nouns as inputs to the multimodal teacher model. Specifically, we examine how varying the proportion of training samples with WordNet-relaxed text embeddings (compared to those using ground truth class names) influences the teacher model's performance and, in turn, the student's performance through Knowledge Distillation (KD). Here, the proportion refers to the ratio of training samples using WordNet-relaxed text embeddings versus those using ground truth class-name-based ones.

\subsubsection{Teacher Performance}
On the CIFAR100 dataset, we conduct a series of experiments where the teacher models $\text{T}_x$ are trained using varying proportions of CLIP WordNet-relaxed text embeddings, ranging from 100\% ground truth class names to 100\% WordNet-relaxed text.
As shown in Figure \ref{fig:wordnet_performance}, higher proportions of WordNet-relaxed text embeddings lead to a more challenging classification task but more robust teacher models, as evidenced by the decreasing teacher validation accuracy.
Notably, the 100\% accuracy for $\text{T}_{x,~100\%\text{gt}~0\%\text{wn}}$ reflects severe label leakage and is misleading, as the teacher model is effectively memorizing the image classes rather than demonstrating true model performance, as discussed further in Section \ref{sec:ablation_noise}.

\subsubsection{Student Performance after KD}
Next, on the CIFAR100 dataset, we distill knowledge from one teacher model $\text{T}_x$ to a student model $\text{S}_s$.
The student results shown in Figure \ref{fig:wordnet_performance} reveal that as the proportion of CLIP WordNet-relaxed text embeddings used by the teacher models $\text{T}_x$ increases, the performance of the student models improves.  
This positive correlation between the proportion of WordNet-relaxed text and the student’s performance supports our hypothesis that using a broader range of semantic information helps the student model generalize better, validating the effectiveness of our WordNet-relaxed regularizer. Notably, the student accuracy with a $\text{T}_{x,~100\%\text{gt}~0\%\text{wn}}$ teacher is the worst, reinforcing our hypothesis that a misleading teacher, despite exhibiting strong performance, transfers little meaningful knowledge to the student, resulting in poor student performance.

These experiments show that while using pure ground truth class names in CLIP text embeddings leads to label leakage and artificially high teacher performance, the introduction of WordNet-relaxed text embeddings provides more robust and semantically diverse representations. As a result, the student model gains more generalized teacher knowledge, improving its classification performance.

\begin{table*}[]
\begin{center}
\begin{small}
\begin{tabular}{lccc}
\hline
\textbf{Teacher Trial} & \textbf{Teacher Val Acc} & \textbf{Student ($\text{T}_{x}\text{S}_s$) Val Acc} & \textbf{Student ($\text{T}_m\text{T}_{x}\text{S}_s$) Val Acc} \\
\hline
$\text{T}_{x,~(\text{img},~0\%\text{gt}~100\%\text{wn})}$ & 64.39\% & \textbf{79.07\%} & \textbf{83.33\%} \\
$\text{T}_{x,~(\text{img},~20\%\text{gt}~80\%\text{wn})}$ & 67.15\%  & 78.98\% & 83.01\% \\
$\text{T}_{x,~(\text{img},~50\%\text{gt}~50\%\text{wn})}$ & 80.84\%  & 78.52\% & 82.87\% \\
$\text{T}_{x,~(\text{img},~80\%\text{gt}~20\%\text{wn})}$ & \textbf{89.66\%}  & 78.40\% & 82.55\% \\
\hline
$\text{T}_{x,~(\text{img},~100\%\text{gt})}$ & 100.00\% & 77.42\% & 82.10\% \\
\hline
$\text{T}_{x,~(\text{img},~80\%\text{gt}~20\%\text{noise})}$ & \textbf{99.10\%} & 77.60\% & 82.49\% \\
$\text{T}_{x,~(\text{img},~50\%\text{gt}~50\%\text{noise})}$ & 98.08\% & 77.50\% & 82.57\% \\
$\text{T}_{x,~(\text{img},~20\%\text{gt}~80\%\text{noise})}$ & 82.29\% & 78.29\% & 82.64\% \\
$\text{T}_{x,~(\text{img},~0\%\text{gt}~100\%\text{noise})}$ & 57.06\% & \textbf{78.50\%} & \textbf{82.99\%} \\
\hline
\end{tabular}
\end{small}
\end{center}
\vskip -0.2in
\caption{Teacher and student top-1 validation accuracy on CIFAR100, where $\text{T}_x$ concatenates CLIP image and text embeddings with varying proportions of \textit{Top}: WordNet-relaxed, \textit{Bottom}: noisy class-name-based prompts. Notably, exact or large ground-truth text proportions give high teacher accuracy but yield lower student accuracy (label leakage). In contrast, increased WordNet relaxation or noise compels $\text{T}_x$ to rely more on image embeddings, ultimately boosting student performance despite lower teacher accuracy.}
\label{tab:TS_concat_clips__ablation_TS_noise}
\vskip -0.1in
\end{table*}

\subsection{Investigation: Concatenating CLIP Image and Text Embeddings}\label{sec:concat_wordnet}

Building on the insights from Section~\ref{sec:wordnettext}, we next investigate the effect of concatenating CLIP image embeddings with CLIP WordNet-relaxed text embeddings for teacher training on the CIFAR100 dataset. Similarly, we vary the proportion of WordNet-relaxed text embeddings in the teacher’s input and analyze both the teacher's performance and its impact on the final student model during KD.

\subsubsection{Teacher Performance}
On the CIFAR100 dataset, we train each teacher model by concatenating CLIP image embeddings with CLIP text embeddings, with a portion of the ground truth class-name text embeddings replaced by WordNet-relaxed ones
(e.g., $\text{T}_{x,~(\text{img},~20\%\text{gt}~80\%\text{wn})}$). Table~\ref{tab:TS_concat_clips__ablation_TS_noise} \textit{Top} outlines the teacher's top-1 validation accuracy for various proportions of WordNet-relaxed text.

From these results, higher WordNet-relaxation proportions reduce the teacher’s accuracy (due to more challenging concatenated embeddings) but increase student accuracy, indicating the effectiveness of such relaxation in KD. Conversely, using exact or large amounts of ground-truth text leads to superficially high teacher accuracy yet ultimately degrades student performance, highlighting label leakage in $\text{T}_{x}$.

\subsubsection{Student Performance after KD}
Next, on the CIFAR100 dataset, we distill knowledge from the multimodal teacher $\text{T}_x$ into the student model $\text{S}_s$. Additionally, we incorporate the unimodal teacher ensemble $\text{T}_m$.
Table~\ref{tab:TS_concat_clips__ablation_TS_noise} \textit{Top} also presents the student’s top-1 validation accuracy with this crossmodal KD.

Compared to the results in Figure \ref{fig:wordnet_performance}, where $\text{T}_{x}$ relies solely on CLIP text embeddings, concatenating CLIP image embeddings with text embeddings leads to higher overall student accuracy. This indicates that incorporating both image and WordNet-relaxed text inputs in a multimodal teacher $\text{T}_{x}$ exploits complementary signals, enhancing the teacher’s semantic diversity and providing robust generalization gains through crossmodal synergy in KD.
Besides, we find that the best performance (83.33\%) is achieved with teacher models $\text{T}_m$ and $\text{T}_{x,~(\text{img},~0\%\text{gt}~100\%\text{wn})}$, outperforming the baseline trials from Section~\ref{sec:final_comparison}, despite the teacher’s moderate validation accuracy. This demonstrates that the semantic richness from full WordNet-based textual inputs offers powerful supervision when combined with other teachers in the KD framework.

\begin{figure*}[h]
    \centering
    \includegraphics[width=0.8\textwidth]{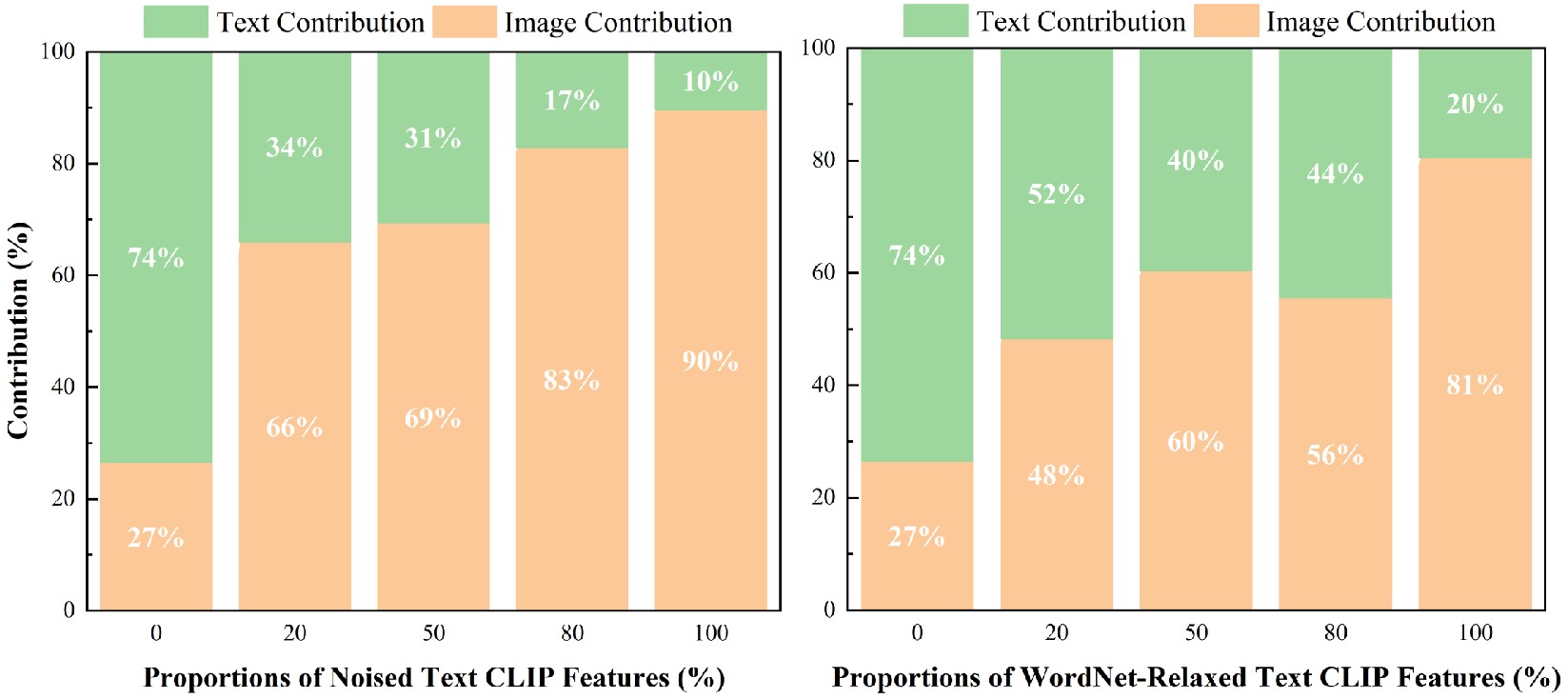}
    \vskip -0.1in
    \caption{Captum feature attribution (text vs.\ image contributions) for teacher $\text{T}_x$ on CIFAR100, with varying proportions of \textit{Left}: CLIP noisy text embeddings and \textit{Right}: CLIP WordNet-relaxed text embeddings. In each set of trials, increasing the noise or WordNet ratio diminishes reliance on direct class tokens, pushing the teacher to depend more on general visual modality features (i.e., CLIP image embeddings) and reducing deceptive ``shortcuts''. Consequently, teachers with 100\% noise or 100\% WordNet text produce the best student accuracy (see Table~\ref{tab:TS_concat_clips__ablation_TS_noise}). Meanwhile, WordNet expansions preserve semantic consistency, enabling the teacher to leverage textual cues more effectively than pure noise, thereby improving crossmodal KD performance.}
    \label{fig:Captum}
    \vskip -0.15in
\end{figure*}

\subsection{Ablation Study}\label{sec:ablation}

\subsubsection{Concatenating CLIP Image and Text Embeddings Under Noise}\label{sec:ablation_noise}

We extend our concatenation approach from Section~\ref{sec:concat_wordnet} by exploring the impact of noisy class-name-based text (obtained by randomly shuffling ground-truth class names) embeddings on teacher and student performance. This is contrasted with our WordNet-relaxed text strategy, serving as an ablation to underscore how semantic coherence influences the teacher’s logits and the final KD outcomes.

\noindent\textbf{Teacher Performance Under Noise:}
Table~\ref{tab:TS_concat_clips__ablation_TS_noise} \textit{Bottom} presents the teacher's top-1 validation accuracy across trials with varying proportions of noisy class names used as text embedding prompts.

Remarkably, nearly every trial, except the one with 100\% noisy texts, reaches high teacher accuracy (yet lower student accuracy), even at noise levels of 80\%. This outcome underscores label leakage, in which the teacher ``memorizes'' answers from text embeddings closely tied to class names. Consequently, the teacher’s validation accuracy no longer indicates genuine generalization but instead reflects the teacher ``cheating'' by exploiting overly specific textual cues.

\noindent\textbf{Student Performance Under Noise}
We then combine these teachers with $\text{T}_m$ to train the student model $\text{S}_s$. Table~\ref{tab:TS_concat_clips__ablation_TS_noise} \textit{Bottom} also presents the student’s top-1 validation accuracy.

Interestingly, increasing the noise proportion in teacher $\text{T}_x$ boosts the student’s accuracy, defying the usual expectation that random label noise would degrade KD. As shown later in Figure~\ref{fig:Captum} \textit{Left}, higher noise forces $\text{T}_x$ to rely more on CLIP image embeddings (i.e., general modality features), thus avoiding label shortcuts and improving student performance. Conversely, lower noise allows $\text{T}_x$ to ``cheat'' by focusing on text embeddings without truly learning the task.
Nevertheless, while noise does mitigate label leakage and reduce a deceptive teacher effect, it also undermines semantic coherence. Consequently, when comparing with the Table~\ref{tab:TS_concat_clips__ablation_TS_noise} \textit{top} trials where WordNet-relaxed text embeddings significantly increased performance (up to 83.33\%), random label noise proves less effective in conveying useful knowledge, ultimately weakening the KD outcome.

In summary, the discrepancy between high teacher accuracy and suboptimal student results highlights that textual label leakage can create illusory teacher performance.
However, the resulting KD signals fail to generalize well to the student. In contrast, replacing ground-truth class names with semantically related words provides genuinely beneficial supervision, as demonstrated in Section~\ref{sec:concat_wordnet}, outperforming noisy approaches and preventing label leakage.

\begin{table}[h]
\begin{center}
\begin{small}
\begin{tabular}{lcccc}
\hline
\multirow{2}{*}{\textbf{Dataset}} 
& \multicolumn{2}{c}{\textbf{$\text{T}_{x}$ Val Acc (\%)}} & \multicolumn{2}{c}{\textbf{$\text{S}_{s}$ Val Acc (\%)}} \\ \cmidrule(lr){2-3} \cmidrule(lr){4-5}
& $\mathbf{n}_\mathrm{pretrained}$ & $\mathbf{n}_\mathrm{relaxed}$ & $\mathbf{n}_\mathrm{pretrained}$ & $\mathbf{n}_\mathrm{relaxed}$ \\
\hline
CIFAR100 & 64.31 & \textbf{64.39} & 82.97 & \textbf{83.33} \\
imb100 & 39.58 & \textbf{40.30} & 50.93 & \textbf{52.58} \\
Scene & 94.66 & \textbf{94.89} & 92.89 & \textbf{93.07} \\
UTKFace & 88.74 & \textbf{89.16} & 86.67 & \textbf{86.71} \\
\hline
\end{tabular}
\end{small}
\end{center}
\vskip -0.2in
\caption{Results for our best-performing trial $\text{T}_m\text{T}_{x,~(\text{img},~0\%\text{gt}~100\%\text{wn})}\text{S}_s$ across multiple datasets, comparing two WordNet-relaxed text embedding strategies: fixed pretrained text embedding $\mathbf{n}_\mathrm{pretrained}$ vs.\ updated text embedding $\mathbf{n}_\mathrm{relaxed}$. ``imb100'' stands for CIFAR100-imb100 dataset.}
\label{tab:ablation_learnable}
\vskip -0.1in
\end{table}

\subsubsection{Text Embedding Updates}
\label{sec:ablation_study2}

Recall that our WordNet-relaxed text embeddings, $\{\mathbf{n}^\star_j\}$, are treated as learnable parameters and updated during teacher training via hierarchical loss and cosine regularization. Here, we investigate whether this learnable embedding strategy benefits both the teacher and student.

With $\text{T}_m\text{T}_{x,~(\text{img},~0\%\text{gt}~100\%\text{wn})}\text{S}_s$ trial, table~\ref{tab:ablation_learnable} summarizes experiments under two approaches: keeping $\mathbf{n}_\mathrm{pretrained}$ fixed vs.\ updating $\mathbf{n}_\mathrm{relaxed}$. Across all datasets, allowing the text embeddings to be updated yields higher validation accuracy for both teacher and student, validating the effectiveness of our learnable embedding strategy.

\subsection{Interpretability}
\label{sec:interpretability_tsne}

In addition to examining teacher and student performance, we further investigate \emph{how} the teacher model combines CLIP image and text embeddings under various class name expansions or noise, using Captum-based feature attribution analysis. %\textcolor{blue}{A t-SNE embedding visualization is also provided in supplementary material Section 2.}

Specifically, on the CIFAR100 dataset, we utilize Captum to measure the relative contribution of each input modality: CLIP image embeddings versus CLIP text embeddings, to the final teacher logits. We compare trials where the teacher is trained on different proportions of noisy or WordNet-relaxed text embedding inputs:
$\text{T}_{x,~(\text{img},~\alpha\%\text{gt}~\beta\%\text{noise})}$ and
$\text{T}_{x,~(\text{img},~\alpha\%\text{gt}~\beta\%\text{wn})}$.
with $\beta \in \{0, 20, 50, 80, 100\}$.

As shown in Figure \ref{fig:Captum} \textit{Left}, increasing the noise proportion from $\text{T}_{x,~(\text{img},~100\%\text{gt}~0\%\text{noise})}$ to $\text{T}_{x,~(\text{img},~0\%\text{gt}~100\%\text{noise})}$ prompts the teacher to rely more on CLIP image embeddings, as reflected in the higher image-component Captum scores. Counterintuitively, injecting random noise into class-name prompts can effectively reduce direct textual shortcuts, yielding a less deceptive teacher by shifting emphasis toward visual features. Consequently, $\text{T}_x$ with 100\% noise produces the highest student accuracy among all noisy trials, as shown in Table \ref{tab:TS_concat_clips__ablation_TS_noise}.

Meanwhile, as shown in Figure \ref{fig:Captum} \textit{Right}, teacher reliance also shifts toward image embeddings when moving from $\text{T}_{x,~(\text{img},~100\%\text{gt}~0\%\text{wn})}$ to $\text{T}_{x,~(\text{img},~0\%\text{gt}~100\%\text{wn})}$. 
Comparing similarly ``difficult'' text embeddings (i.e., the same proportion of noise or WordNet) in the \textit{Left} (noisy) versus \textit{Right} (WordNet-relaxed) plots reveals that the WordNet-relaxed teacher leverages textual cues more effectively, as demonstrated by stronger text-component Captum scores. By preserving semantic consistency with WordNet expansions yet eliminating exact class tokens, our method genuinely integrates additional textual signals alongside the visual modality, enhancing crossmodal KD. Consequently, $\text{T}_x$ with 100\% WordNet delivers the highest student accuracy among all KD trials, as shown in Table \ref{tab:TS_concat_clips__ablation_TS_noise}.

In summary, these findings confirm our hypothesis in Section~\ref{sec:problemDef} that exact class names cause label leakage and text-driven classification, whereas noisy or WordNet-relaxed prompts mitigate textual memorization and shift the model’s reliance toward general visual modality features (i.e., CLIP image embeddings). Furthermore, WordNet expansions provide richer semantic breadth than simple noise, ultimately yielding a more robust multimodal representation.

\section{Conclusion}\label{sec:conclusion}

We propose a multi-teacher crossmodal KD framework that employs CLIP image embeddings and learnable WordNet-relaxed text embeddings in a multimodal teacher to distill knowledge into a unimodal student. By replacing exact class names with semantically richer nouns from WordNet, we address label leakage and deceptive teacher behavior, which artificially boosts teacher accuracy without improving student learning.
Extensive experiments show consistent gains in student performance when teachers adopt WordNet-relaxed text prompts, demonstrating that broader semantic cues guide the student toward more robust features. Interpretability analyses reveal that exact class-name prompts cause over-reliance on text embeddings, while entirely noisy labels fail to leverage text effectively. In contrast, WordNet expansions prompt the teacher to better integrate visual and textual information, ultimately enhancing crossmodal KD.

\end{document}